# On Axiomatization of Probabilistic Conditional Independencies

S.K.M. Wong and Z.W. Wang
Department of Computer Science
University of Regina
Regina, Saskatchewan
Canada S4S 0A2
wong@cs.uregina.ca, fax:(306)585-4745

## Abstract

This paper studies the connection between probabilistic conditional independence in uncertain reasoning and data dependency in relational databases. As a demonstration of the usefulness of this preliminary investigation, an alternate proof is presented for refuting the conjecture suggested by Pearl and Paz that probabilistic conditional independencies have a complete axiomatization.

## 1 Introduction

This paper studies the relationship between the notion of dependencies in probability theory and that of data dependencies in relational database theory. In the relational database model, many types of constraints called dependencies are used as a semantic language to express properties of data. Data dependencies, such as multivalued dependency, play an important role in database schema design. They are used to decompose a relation into a number of smaller relations in order to reduce data redundancy and avoid anomalies when updating relations. Thus, data dependencies have been studied extensively in the standard relational model for databases [Beeri et al. 1983].

An important and useful tool in investigating the properties of a class of dependencies is a complete set of inference rules (axioms), usually called a complete axiomatization [Parker and Parsaye-Ghomi 1980; Maier 1983]. Completeness means that any dependency that is logically implied by a set $\Sigma$ of dependencies can be derived from $\Sigma$ by repeated applications of the axioms. One may also use these qualitative axioms to compare the expressive power of different knowledge representations, and derive interesting and powerful theorems that may not be easily obtained from the numerical representation of data.

Since a great deal about data dependencies in the standard relational model is known, our objective in this paper is to establish a link between probabilistic dependencies and data dependencies. To demonstrate the significance of this investigation, we will use this approach to provide an alternate proof [Studeny 1990a, 1990b, 1989] for refuting the conjecture suggested by Pearl and Paz (1985) that probabilistic conditional independencies have a complete axiomatization [Geiger and Pearl 1988]. More importantly, this preliminary study gives us a deeper insight into the algebraic structure of conditional independence, and may lead to a complete axiomatization of a subclass of probabilistic dependencies.

This paper is organized as follows. In Section 2, we introduce the notion of generalized embedded multivalued dependency (GEMVD) in an extended relational model. We show in Section 3 that this class of data dependencies is equivalent to the notion of probabilistic conditional independence. In Section 4, we demonstrate that embedded multivalued dependency in the conventional relational model, a subclass of GEMVD, has no complete axiomatization. Consequently, no formal theory with a finite number of inference axioms can be complete for probabilistic conditional independencies.

## 2 Basic Notions in Relational Models

### 2.1 The standard relational database model

In the relational database model [Maier 1983], the data are viewed as finite tables called relations. The columns of a relation correspond to attributes, and the rows to records or tuples. Each attribute has an associated domain of values. A tuple can be interpreted as a mapping from the attributes to their respective domains. We say that a relation $r$ is a relation over a set of attributes $X$, if the columns of $r$ correspond precisely to those attributes in $X$.

Let $N$ be a finite set of attributes (variables). We will use the letters $A, B, C, \ldots$ to denote single attributes,

and the letters $..., X, Y, Z$ to denote sets of attributes. Suppose $r$ is a relation over a set of attributes $X$, $t$ is a tuple in $r$, and $A$ is an attribute in $X$. The tuple $t$ maps the attribute $A$ to $t(A)$, and $t(A)$ is called the $A$-value of $t$. If $Y$ is a subset of $X$, then $t(Y)$ is a tuple defined only on those attributes in $Y$. It is understood that $t$ maps each attribute $A$ in $Y$ to $t(A)$. We call $t(Y)$ a $Y$-value in $r$, which is also referred to as a configuration of $Y$ in $r$. The projection of the relation $r$ onto $Y$, written $r(Y)$, is obtained by removing the columns of $r$ that do not correspond to those attributes in $Y$ and identifying equal tuples, i.e., $r(Y) = \{t(Y) \mid t \text{ is a tuple of } r\}$. Consider two relations $r(R_1)$ and $r(R_2)$. Let $R = R_1 R_2$ (where the union of two sets $R_1$ and $R_2$ is written as $R_1 R_2$). The natural join of $r_1$ and $r_2$, written $r_1 \bowtie r_2$, is the relation $r(R)$ of all tuples $t$ over $R$ such that there are tuples $t_{r_1} \in r_1$ and $t_{r_2} \in r_2$ with $t_{r_1} = t(R_1)$ and $t_{r_2} = t(R_2)$.

The data of interest satisfy certain constraints. These constraints are usually called dependencies in the relational database model. In this paper, we only consider multivalued dependencies (MVDs). A relation $r$ over $R$ satisfies the $MVD\ X \twoheadrightarrow Y|Z$ if for all tuples $t_1$ and $t_2$ in $r$, if $t_1(X) = t_2(X)$, then there is a tuple $t$ in $r$ such that

$$t(XY) = t_2(XY) \text{ and } t(Z) = t_2(Z),$$

where $X$ and $Y$ are disjoint subsets of attributes and $Z = R - XY$.

We can define the MVD $X \twoheadrightarrow Y$ in another way. Let $x$ be an $X$-value in a relation $r$, and define $Z_r(x)$ to be the set of all $Z$-values $z$ associated with the $X$-value $x$, namely:

$$Z_r(x) = \{z \mid \text{there is a tuple } t \text{ in } r$$
$$\text{such that } t(Z) = z \text{ and } t(X) = x\}.$$

For convenience, we will write $Z_r(x)$ as $Z(x)$ if no confusion arises.

**Lemma 1** [Sagiv and Walecka 1982] *The MVD $X \twoheadrightarrow Y|Z$ holds in a relation $r$ over $XYZ$, if and only if $Z_r(x) = Z_r(xy)$ for all $XY$-values $xy$ in $r$.*

MVDs can also be equivalently defined by the following lemma.

**Lemma 2** *The MVD $X \twoheadrightarrow Y|Z$ holds in a relation $r$ over $XYZ$, if and only if*

$$|YZ(x)| = |Z(xy)| \cdot |Y(xz)|.$$

Under certain circumstances, it is necessary to consider constraints that hold in the projection of a relation onto a subset of its attributes but do not necessarily hold in the entire relation. Constraints of this type are called *embedded*. In particular, the *embedded multivalued dependency* (EMVD) $X \twoheadrightarrow Y|Z$ holds in a relation $r(R)$ over a set of attributes $R$ such that $XYZ \subseteq R$, if the MVD $X \twoheadrightarrow Y|Z$ holds in $r(XYZ)$ (i.e., in the projection of $r$ onto $XYZ$).

## 2.2 An extended relational model

In our extended relational model [Wong et al. 1994], each relation $\Phi_R$ is defined by a real-valued function $\phi_R$ on a set of attributes $R = \{A_1, A_2, ..., A_m\}$ as shown in Figure 1, where $t_i = (t_{i1}, t_{i2}, ..., t_{im})$ is a tuple of $R$. It is important to note that the function $\phi_R$ defines the values of the attribute $f_{\phi_R}$ in relation $\Phi_R$. In the conventional relational database model, one may view $\phi_R$ as a *constant* function. In this case, it may not be necessary to use such a function to define a relation.

$$\Phi_R = \begin{array}{|cccc|c|} \hline A_1 & A_2 & ... & A_m & f_{\phi_R} \\ \hline t_{11} & t_{12} & ... & t_{1m} & \phi_R(t_1) \\ t_{21} & t_{22} & ... & t_{2m} & \phi_R(t_2) \\ \vdots & \vdots & & \vdots & \vdots \\ t_{s1} & t_{s2} & ... & t_{sm} & \phi_R(t_s) \\ \hline \end{array}$$

Figure 1: A relation in the extended relational database model.

If $\phi_R(t_i) \neq 0$ for each $t_i (1 \leq i \leq s)$, we can define the *inverse* relation $\Phi_R^{-1}$ for $\Phi_R$ as:

$$\Phi_R^{-1} = \begin{array}{|cccc|c|} \hline A_1 & A_2 & ... & A_m & f_{\phi_R^{-1}} \\ \hline t_{11} & t_{12} & ... & t_{1m} & 1/\phi_R(t_1) \\ t_{21} & t_{22} & ... & t_{2m} & 1/\phi_R(t_2) \\ \vdots & \vdots & & \vdots & \vdots \\ t_{s1} & t_{s2} & ... & t_{sm} & 1/\phi_R(t_s) \\ \hline \end{array}$$

Figure 2: The inverse relation $\Phi_R^{-1}$ of $\Phi_R$.

In addition to the select, project, and natural join operators in the standard relational database model, we introduce here two new relational operators called *marginalization* and *product join*.

(i) *Marginalization*

Let $X$ be a subset of attributes of $R$. The marginalization of $\Phi_R$ on the subset of attributes $X \cup \{f_{\phi_R}\}$,

written $\Phi_R^{\downarrow X}$, is a relation on $X \cup \{f_{\phi_R}\}$. We can construct the *marginal* $\Phi_R^{\downarrow X}$ of $\Phi_R$ as follows:

1. First project the relation $\Phi_R$ on $X \cup \{f_{\phi_R}\}$, without eliminating identical tuples (configurations).

2. For every configuration $t_X$ in the relation obtained from Step 1, replace the set of tuples with the same $X$-value $t_X$ by the tuple:

$$t_X * (\sum_{t_{R-X}} \phi_R(t_X * t_{R-X})),$$

where $t_{R-X} = t(R-X), t = t_X * t_{R-X}$, and $t$ is a tuple in $\Phi_R$. The symbol $*$ denotes the concatenation of two tuples.

Consider, for example, the relation $\Phi_X$ in Figure 3, defined by a function $\phi_X$ on $X = \{A_1, A_2, A_3\}$.

$$\Phi_X = \begin{array}{|ccc|l|} \hline A_1 & A_2 & A_3 & f_{\phi_X} \\ \hline 0 & 0 & 0 & \phi_X(0,0,0) = d_1 \\ 0 & 0 & 1 & \phi_X(0,0,1) = d_2 \\ 0 & 1 & 0 & \phi_X(0,1,0) = d_3 \\ 0 & 1 & 1 & \phi_X(0,1,1) = d_3 \\ 1 & 0 & 0 & \phi_X(1,0,0) = d_4 \\ 1 & 0 & 1 & \phi_X(1,0,1) = d_4 \\ 1 & 1 & 0 & \phi_X(1,1,0) = d_5 \\ 1 & 1 & 1 & \phi_X(1,1,1) = d_6 \\ \hline \end{array}$$

Figure 3: A relation $\Phi_X$ with $X = \{A_1, A_2, A_3\}$ defined by a function $\phi_X$.

Suppose we want to compute the marginal $\Phi_X^{\downarrow A_1 A_2}$. From Step 1, we obtain the table in Figure 4 by projecting $\Phi_X$ on $\{A_1, A_2, f_{\phi_X}\}$ without eliminating identical tuples.

$$\Phi_X(\{A_1, A_2, f_{\phi_X}\}) = \begin{array}{|cc|c|} \hline A_1 & A_2 & f_{\phi_X} \\ \hline 0 & 0 & d_1 \\ 0 & 0 & d_2 \\ 0 & 1 & d_3 \\ 0 & 1 & d_3 \\ 1 & 0 & d_4 \\ 1 & 0 & d_4 \\ 1 & 1 & d_5 \\ 1 & 1 & d_6 \\ \hline \end{array}$$

Figure 4: The "projection" of $\Phi_X$ on $\{A_1, A_2, f_{\phi_X}\}$.

The resultant relation, i.e., the marginal $\Phi_X^{\downarrow A_1 A_2}$, constructed from Step 2 is shown in Figure 5.

$$\Phi_X^{\downarrow A_1 A_2} = \begin{array}{|cc|c|} \hline A_1 & A_2 & f_{\phi_X} \\ \hline 0 & 0 & d_1 + d_2 \\ 0 & 1 & d_3 + d_3 \\ 1 & 0 & d_4 + d_4 \\ 1 & 1 & d_5 + d_6 \\ \hline \end{array}$$

Figure 5: The marginal $\Phi_X^{\downarrow A_1 A_2}$ of relation $\Phi_X$.

(ii) *Product Join*

Consider two relations $\Phi_X$ and $\Psi_Y$ defined respectively by the functions $\phi_X$ and $\psi_Y$. The *product join* of $\Phi_X$ and $\Psi_Y$, written $\Phi_X \times \Psi_Y$, is defined as follows:

1. First form the natural join, $\Phi_X \bowtie \Psi_Y$, of the two relations $\Phi_X$ and $\Psi_Y$.

2. Add a new column labeled by the attribute $f_{\phi_X \cdot \psi_Y}$ to the resultant relation $\Phi_X \bowtie \Psi_Y$. The values of $f_{\phi_X \cdot \psi_Y}$ are defined by the product $\phi_X(t(X)) \cdot \psi_Y(t(Y))$, where $t$ is a tuple of $XY$ such that $t(X) = t_X \in \Phi_X(X)$ and $t(Y) = t_Y \in \Psi_Y(Y)$.

3. The product join $\Phi_X \times \Psi_Y$ is obtained by projecting the relation constructed in Step 2 on the set of attributes $XY \cup \{f_{\phi_X \cdot \psi_Y}\}$.

An example of the product join operation is illustrated in Figure 6.

$$\begin{array}{|cc|c|} \hline A_1 & A_2 & f_{\phi_X} \\ \hline 0 & 0 & a_1 \\ 0 & 1 & a_2 \\ 1 & 0 & a_3 \\ 1 & 1 & a_4 \\ \hline \end{array} \overset{(1)}{\bowtie} \begin{array}{|cc|c|} \hline A_1 & A_2 & f_{\psi_Y} \\ \hline 0 & 0 & b_1 \\ 0 & 1 & b_2 \\ 1 & 0 & b_3 \\ 1 & 1 & b_4 \\ \hline \end{array}$$

$$= \begin{array}{|ccc|cc|} \hline A_1 & A_2 & A_3 & f_{\phi_X} & f_{\psi_Y} \\ \hline 0 & 0 & 0 & a_1 & b_1 \\ 0 & 0 & 1 & a_1 & b_2 \\ 0 & 1 & 0 & a_2 & b_3 \\ 0 & 1 & 1 & a_2 & b_4 \\ 1 & 0 & 0 & a_3 & b_1 \\ 1 & 0 & 1 & a_3 & b_2 \\ 1 & 1 & 0 & a_4 & b_3 \\ 1 & 1 & 1 & a_4 & b_4 \\ \hline \end{array} \overset{(2)}{\longrightarrow}$$

$$\begin{array}{|ccc|ccc|} \hline A_1 & A_2 & A_3 & f_{\phi_X} & f_{\psi_Y} & f_{\phi_X \cdot \psi_Y} \\ \hline 0 & 0 & 0 & a_1 & b_1 & a_1 \cdot b_1 \\ 0 & 0 & 1 & a_1 & b_2 & a_1 \cdot b_2 \\ 0 & 1 & 0 & a_2 & b_3 & a_2 \cdot b_3 \\ 0 & 1 & 1 & a_2 & b_4 & a_2 \cdot b_4 \\ 1 & 0 & 0 & a_3 & b_1 & a_3 \cdot b_1 \\ 1 & 0 & 1 & a_3 & b_2 & a_3 \cdot b_2 \\ 1 & 1 & 0 & a_4 & b_3 & a_4 \cdot b_3 \\ 1 & 1 & 1 & a_4 & b_4 & a_4 \cdot b_4 \\ \hline \end{array} \overset{(3)}{\longrightarrow}$$

| $A_1$ | $A_2$ | $A_3$ | $f_{\phi_X \cdot \psi_Y}$ |
|---|---|---|---|
| 0 | 0 | 0 | $a_1 \cdot b_1$ |
| 0 | 0 | 1 | $a_1 \cdot b_2$ |
| 0 | 1 | 0 | $a_2 \cdot b_3$ |
| 0 | 1 | 1 | $a_2 \cdot b_4$ |
| 1 | 0 | 0 | $a_3 \cdot b_1$ |
| 1 | 0 | 1 | $a_3 \cdot b_2$ |
| 1 | 1 | 0 | $a_4 \cdot b_3$ |
| 1 | 1 | 1 | $a_4 \cdot b_4$ |

$= \Phi_X \times \Psi_Y$.

Figure 6: The product join $\Phi_X \times \Psi_Y$ of relations $\Phi_X$ and $\Psi_Y$.

### 2.3 Generalized Multivalued Dependency

Here we introduce the key notion of *generalized multivalued dependency* (GMVD) in the extended relational model [Wong et al. 1994].

let $\Phi_R$ be a relation over the set of attributes $R \cup \{f_{\phi_R}\}$ as shown in Figure 1. Let $X$ and $Y$ be disjoint subsets of $R$ and $Z = R - XY$. A relation $\Phi_R$ satisfied the GMVD $X \multimap\!\!\!\rightarrow Y|Z$ if $\Phi_R$ *decomposes losslessly* into relations $\Phi_R^{\downarrow XY}$ and $\Phi_R^{\downarrow XZ}$, that is:

$$\Phi_R = \Phi_R^{\downarrow XY} \otimes \Phi_R^{\downarrow XZ}.$$

The above *monotone join operator* $\otimes$ is defined as: for any $V, W \subseteq R$,

$$\Phi_R^{\downarrow V} \otimes \Phi_R^{\downarrow W} = \Phi_R^{\downarrow V} \times \Phi_R^{\downarrow W} \times (\Phi_R^{\downarrow V \cap W})^{-1},$$

where $\times$ is the product join operator and $(\Phi_R^{\downarrow V \cap W})^{-1}$ is the inverse relation of $\Phi_R^{\downarrow V \cap W}$.

Similar to the standard relational database model, we say that the *generalized embedded multivalued dependency* (GEMVD) $X \multimap\!\!\!\rightarrow Y|Z$ holds in a relation $\Phi_R$ such that $XYZ \subseteq R$, if the GMVD $X \multimap\!\!\!\rightarrow Y|Z$ holds in $\Phi_R^{\downarrow XYZ}$ (i.e., in the marginal $\Phi_R^{\downarrow XYZ}$ of $\Phi_R$).

Now we want to show that multivalued dependency is a subclass of generalized multivalued dependency.

**Theorem 1** *Let $\Phi_R$ be a constant relation over $R \cup \{f_{\phi_R}\}$, and let $X$ and $Y$ be disjoint subsets of $R$ and $Z = R - XY$. Relation $\Phi_R$ satisfies the GMVD $X \multimap\!\!\!\rightarrow Y|Z$, i.e., $\Phi_R = \Phi_R^{\downarrow XY} \otimes \Phi_R^{\downarrow XZ}$, if and only if the relation $\hat{\Phi}_R = \Phi_R(R)$ over $R$ satisfies the MVD $X \rightarrow\!\!\!\rightarrow Y|Z$.*

**Proof:** Let $\Phi_R$ be a relation defined by the function $\phi_R$ on $R$. By the definition of marginalization, the marginal $\Phi_R^{\downarrow XY}$ is a relation over the set of attributes $XY \cup \{f_{\phi_{XY}}\}$ defined by the function $\phi_{XY}$, namely: for any tuple $xy = t_{XY} = t(XY)$ and $t \in \hat{\Phi}_R$,

$$\begin{aligned}\phi_{XY}(xy) &= \phi_{XY}(t_{XY}) \\ &= \sum_{t_Z} \phi_R(t_{XY} * t_Z) \\ &= \sum_z \phi_R(xyz),\end{aligned}$$

where $z = t_Z = t(Z)$. By assumption, $\phi_R$ is a constant relation, i.e., $\phi_R(t) = c$ for any tuple $t \in \hat{\Phi}_R$. It immediately follows:

$$\phi_{XY}(xy) = \sum_z \phi_R(xyz) = c|Z(xy)|,$$

where $|Z(xy)|$ is the number of distinct $Z$-values for a given $XY$-value $xy$ in $\hat{\Phi}_R$. Similarly, from the marginal $\phi_R^{\downarrow XZ}$, we obtain:

$$\phi_{XZ}(xz) = \sum_y \phi_R(xyz) = c|Y(xz)|,$$

where $|Y(xz)|$ is the number of distinct $Y$-values for a given $XZ$-value $xz$ in $\hat{\Phi}_R$.

By definition, the relation $\Phi_R^{\downarrow XY} \otimes \Phi_R^{\downarrow XZ}$ is defined by the function $\rho_R$ on $R = XYZ$: for any tuple $xyz \in \Phi_R(XY) \bowtie \Phi_R(XZ)$,

$$\rho_R(xyz) = \frac{\phi_{XY}(xy) \cdot \phi_{XZ}(xz)}{\phi_X(x)}.$$

The function $\phi_X$ is defined by:

$$\phi_X(x) = \sum_{t_{YZ}} \phi_R(t_X * t_{YZ}) = \sum_{yz} \phi_R(xyz) = c|YZ(x)|,$$

where $|YZ(x)|$ is the number of distinct tuples for a given $X$-value in $\hat{\Phi}_R$.

Thus, the function $\rho_R(xyz)$ can be expressed as:

$$\rho_R(xyz) = \frac{c|Z(xy)| \cdot c|Y(xz)|}{c|YZ(x)|}.$$

On the other hand, relation $\Phi_R$ is defined by the constant function $\phi_R$ on $R$, i.e., for any tuple $xyz = t \in \hat{\Phi}_R$, $\phi_R(xyz) = c$. Clearly, the condition $\Phi_R = \Phi_R^{\downarrow XY} \otimes \Phi_R^{\downarrow XZ}$ is satisfied if and only if $\rho_R = \phi_R$. Therefore, the condition $\rho_R = \phi_R$ holds if and only if for any $XYZ$-value $xyz$ in $\hat{\Phi}_R$,

$$|YZ(x)| = |Z(xy)| \cdot |Y(xz)|.$$

By Lemma 2, this equality holds, if and only if relation $\hat{\Phi}_R$ satisfies the MVD $X \rightarrow\!\!\!\rightarrow Y|Z$.    □

By Theorem 1, embedded multivalued dependency is obviously a subclass of generalized embedded multivalued dependency.

**Corollary 1** *The constant relation $\Phi_{XYZ} = \Phi_R^{\downarrow XYZ}$ satisfies the GEMVD $X \multimap\!\!\!\rightarrow Y|Z$ such that $XYZ \subseteq R$, if and only if the relation $\hat{\Phi}_{XYZ} = \Phi_{XYZ}(XYZ)$ satisfies the EMVD $X \rightarrow\!\!\!\rightarrow Y|Z$.*

## 3 Probabilistic Conditional Independence

We will show in this section that the notion of probabilistic conditional independence is equivalent to that of generalized multivalued dependency.

Given a joint probabilistic distribution $\phi_R$ on a set of variables (attributes) $R$, one can construct another function called a *marginal distribution* $\phi_X$ on a subset $X$ of $R$ [Pearl 1988]. Let $t$ denote a configuration (tuple) of $R$ and $t_X = t(X)$ be a configuration of $X \subseteq R$. The marginal $\phi_X$ of $\phi_R$ is defined by: for any configuration $t_X$,

$$\phi_X(t_X) = \sum_{t_{R-X}} \phi_R(t_X * t_{R-X}),$$

where $t_{R-X} = t(R-X)$ and $t = t_X * t_{R-X}$.

Let $X$ and $Y$ be disjoint subsets of $R$ and $Z = R - XY$. We say that $Y$ and $Z$ are *conditionally independent given* $X$, if for any $XYZ$-value $xyz$,

$$\phi_{XYZ}(xyz) = \frac{\phi_{XY}(xy) \cdot \phi_{XZ}(xz)}{\phi_X(x)}.$$

Recall that any real-valued function can be represented by a relation in the extended relational database model. Thus, a joint probability distribution $\phi_R$ can be conveniently represented as a relation $\Phi_R$ (see Figure 1). Furthermore, by the definition of marginalization, the marginal $\Phi_R^{\downarrow X}$ of the relation $\Phi_R$ represents the marginal $\phi_X$ of the joint distribution $\phi_R$. It immediately follows that probabilistic conditional independence can be equivalently stated as a generalized multivalued dependency, namely:

$$\begin{aligned} \Phi_R &= \Phi_R^{\downarrow XY} \times \Phi_R^{\downarrow XZ} \times (\Phi_R^{\downarrow X})^{-1} \\ &= \Phi_R^{\downarrow XY} \otimes \Phi_R^{\downarrow XZ}, \end{aligned}$$

where the marginals $\Phi_R^{\downarrow XY}, \Phi_R^{\downarrow XZ}$, and $\Phi_R^{\downarrow X}$ of $\Phi_R$ represent the marginal distributions $\phi_{XY}$, $\phi_{XZ}$, and $\phi_X$ of $\phi_R$, respectively. These results are summarized in the following theorem.

**Theorem 2** *Let $\phi_R$ be a joint probability distribution on a set of variables $R$, and let $X$ and $Y$ be disjoint subsets of $R$ and $Z = R - XY$. The sets of variables $Y$ and $Z$ are conditionally independent given $X$, if and only if the relation $\Phi_R$ defined by $\phi_R$ satisfies the generalized multivalued dependency $X \multimap Y | Z$.*

## 4 Axiomatization of Embedded Multivalued Dependencies

Based on the results in Sections 2 and 3, it is clear that the class of constraints referred to as embedded multivalued dependency (EMVD) in the standard relational database model is a subclass of probabilistic conditional independence.

In this section, we outline the proof that a subclass of EMVDs, called Z-EMVD, does *not* have a complete axiomatization. (For a more detailed discussion, see the paper by Sagiv and Walecka (1982)). As a result, there is no *complete formal theory* for probabilistic conditional independencies. In other words, the conjecture of Pearl and Paz (1985) is being refuted.

A Z-EMVD is a EMVD of the form $X \twoheadrightarrow Y|Z$, where $Z$ is a fixed set of attributes, and $X, Y, Z$ are pairwise disjoint. Let $\Sigma$ be a set of Z-EMVDs. We can construct a directed graph $G_\Sigma$. Every node $[X]$ in $G_\Sigma$ represents a subset $X$ of attributes that is disjoint from $Z$. There is a arc in $G_\Sigma$ from $[X]$ to $[Y]$ if either:

(a) $Y \subseteq X$, i.e., $X \twoheadrightarrow Y|Z$ is a trivial Z-EMVD, or

(b) $X \twoheadrightarrow Y|Z$ is a nontrivial Z-EMVD in $\Sigma$.

Note that $G_\Sigma$ contains only those nodes which are part of an arc.

Since MVDs are transitive, by definition, $X \twoheadrightarrow Y|Z$ is implied by $\Sigma$ if there is a directed path from $[X]$ to $[XY]$. Thus, we call the set Z-EMVD$^C(\Sigma)$ defined by:

$$\begin{aligned} &\text{Z-EMVD}^C(\Sigma) \\ &= \{X \twoheadrightarrow Y|Z \mid \text{ there is a directed path} \\ &\quad \text{from } [X] \text{ to } [XY] \text{ in } G_\Sigma\}, \end{aligned}$$

the *cover* of $\Sigma$.

**Lemma 3** [Sagiv and Walecka 1982] *A non-trivial EMVD $W \twoheadrightarrow V_1|V_2$ is implied by a set $\Sigma$ of Z-EMVDs, if and only if there is a Z-EMVD $\sigma$ in Z-EMVD$^C(\Sigma)$ such that $W \twoheadrightarrow V_1|V_2$ can be derived from $\sigma$ by the respective symmetry, augmentation, and projection inference axioms:*

(i)   $X \twoheadrightarrow Y|Z \implies X \twoheadrightarrow Z|Y$,
(ii)  $X \twoheadrightarrow Y|ZW \implies XW \twoheadrightarrow Y|Z$,
(iii) $X \twoheadrightarrow Y|Z,\ Y' \subseteq Y,\ \text{and}\ Z' \subseteq Z \implies$
      $X \twoheadrightarrow X'|Z'$.

The detailed proof of this lemma is given by Sagiv and Walecka (1982).

Given any positive integer $n$, we can always construct a relation $r$ satisfying the following set $\Sigma^{(n)}$ of Z-MVDs:

$$\begin{aligned} X_0 &\twoheadrightarrow X_1|Z, \\ X_1 &\twoheadrightarrow X_2|Z, \\ &\vdots \\ X_{n-2} &\twoheadrightarrow X_{n-1}|Z, \\ X_{n-1} &\twoheadrightarrow X_0|Z, \end{aligned}$$

where $X_0, X_1, ..., X_{n-1}$ and $Z$ are pairwise disjoint subsets of attributes. That is, $\Sigma^{(n)}$ contains the Z-MVDs $X_i \twoheadrightarrow X_{i+1}|Z$, for all $0 \leq i \leq n-2$, and $X_{n-1} \twoheadrightarrow X_0|Z$. It is understood that addition and subtraction of indices are done modulo $n$. For example, $X_n$ is $X_0$, and $X_{-1}$ is $X_{n-1}$.

We will use $\Sigma^{(n)}$ as a counter example to prove that embedded multivalued dependencies do not have a complete axiomatization. The set $\Sigma^{(n)}$ defined above satisfies the following two properties.

*Property 1*   *The Z-EMVD $X_0 \twoheadrightarrow X_{n-1}|Z$ is implied by $\Sigma^{(n)}$.*

This property can be illustrated by an example with $n = 4$. The directed graph $G_{\Sigma^{(4)}}$ is shown in Figure 7, which has the following nodes: $[X_i]$ for $0 \leq i \leq 3$, and $[X_iX_{i+1}]$ for $0 \leq i \leq 3$, The arcs corresponding to trivial dependencies are denoted by solid arrows, and the arcs corresponding to the dependencies in $\Sigma^{(4)}$ are denoted by broken arrows. Clearly, there is path from $[X_0]$ to $[X_3X_0]$. Hence, $X_0 \twoheadrightarrow X_3|Z$ is implied by $\Sigma^{(4)}$. The above argument can be easily extended to any positive integer $n$.

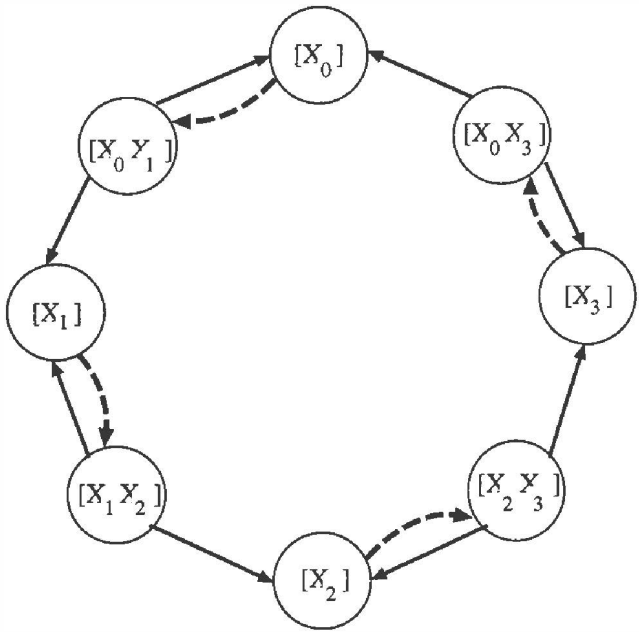

Figure 7: The directed graph $G_{\Sigma^{(4)}}$.

*Property 2*   Let $\Sigma'$ be a subset of $n - 1$ dependencies from $\Sigma^{(n)}$. If $\sigma'$ is a nontrivial EMVD implied by $\Sigma'$, then there is a Z-EMVD $\sigma$ in $\Sigma'$ such that $\sigma'$ can be derived from the symmetry, augmentation, and projection inference axioms.

Consider again the graph $G_\Sigma^{(4)}$ in Figure 7. Obviously, a path in $G_\Sigma^{(4)}$ that corresponds to a Z-EMVD implied by $\Sigma^{(4)}$ must start at either $[X_i]$ or $[X_{i+1}]$ and terminate at $[X_iX_{i+1}]$, for some $0 \leq i \leq 4$. Also, there is an arc from $[X_i]$ to $[X_iX_{i+1}]$, because $X_i \twoheadrightarrow X_{i+1}|Z$ is in $\Sigma^{(4)}$. It is not difficult to see that there is a path from $[X_{i+1}]$ to $[X_iX_{i+1}]$ for all $i$, namely, $X_{i+1} \twoheadrightarrow X_i|Z$ is implied by $\Sigma^{(4)}$, $0 \leq i \leq 4$. However, every path from $[X_{i+1}]$ to $[X_iX_{i+1}]$ uses all the arcs from $[X_i]$ to $[X_iX_{i+1}]$, (i.e., all the arcs that correspond to the Z-EMVDs in $\Sigma^{(4)}$). Therefore, if $\Sigma'$ is obtained by removing one of the Z-EMVDs in $\Sigma^{(4)}$, then none of the Z-EMVDs $X_{i+1} \twoheadrightarrow X_i|Z$, $0 \leq i \leq 4$, is in the cover Z-EMVD$^C(\Sigma')$. As a result, the only Z-EMVDs in Z-EMVD$^C(\Sigma')$ are those in $\Sigma'$. Thus, If $\sigma'$ is a nontrivial EMVD implied by $\Sigma'$, then there is a Z-EMVD $\sigma$ in $\Sigma'$ such that $\sigma'$ can be derived from the symmetry, augmentation, and projection inference axioms.

**Theorem 3** [Sagiv and Walecka 1982]   *Embedded multivalued dependencies do not have a complete axiomatization.*

**Proof:**   We will prove this theorem by contradiction. Suppose that EMVDs have a complete axiomatization with a *finite* number of inference axioms, say $m$. Let $n \geq m$, and consider the set $\Sigma^{(n)}$ of Z-EMVDs defined earlier.

We claim: "Suppose $\tau$ is a nontrivial EMVD that can be derived from $\Sigma^{(n)}$ by the inference axioms in the given complete formal theory for EMVDs. Then $\tau$ can be derived from one of the Z-MVDs in $\Sigma^{(n)}$ by the symmetry, augmentation, and projection inference axioms."

This claim can be proved by induction on the number of applications of inference axioms in the derivation of $\tau$ from $\Sigma^{(n)}$.

*Basis.* Zero applications. The EMVD is one of the EMVDs in $\Sigma^{(n)}$, and hence the claim is trivially true.

*Induction.* Let $\sigma_1, \sigma_2, ..., \sigma_r$ be a derivation of $\tau$ from $\Sigma^{(n)}$ by $k$ applications of the inference axioms. There is a positive integer $p$ such that $\tau$ is a direct consequence of the EMVDs $\sigma_{i_1}, \sigma_{i_2}, ..., \sigma_{i_p}$ by one of the inference axioms. Each $\sigma_{i_j}$ is either a trivial EMVD, an EMVD in $\Sigma^{(n)}$, or can be derived from $\Sigma^{(n)}$ by fewer than $k$ applications of the inference axioms. By the inductive hypothesis, every nontrivial EMVD that can be derived from $\Sigma^{(n)}$ by fewer than $k$ applications of the inference axioms can also be derived from one of the EMVDs in $\Sigma^{(n)}$ by augmentation, projection, and complementation. Thus each EMVD $\sigma_{i_j}$ is either trivial or implied by a single EMVD in $\Sigma^{(n)}$. Since $p < m \leq n$, this means that $\tau$ is implied by fewer than $n$ EMVDs in $\Sigma^{(n)}$. By Property 2, we can immediately conclude that $\tau$ can be derived by augmentation, projection, and complementation from one of the EMVDs in $\Sigma^{(n)}$. This completes the proof of the claim.

However, by Property 1, $\Sigma^{(n)}$ implies the EMVD $X_0 \twoheadrightarrow X_{n-1}|Z$, and this nontrivial EMVD cannot be derived by the symmetry, augmentation, and projection inference axioms from *any* EMVD in $\Sigma^{(n)}$. This

observation is in contradiction with the above claim. Therefore, the given formal theory cannot be complete for EMVDs. □

## 5 Conclusions

In this preliminary report, we have shown explicitly the connection between probabilistic conditional independence and generalized embedded multivalued dependency in our extended relational model for databases. We have demonstrated the usefulness of this linkage by presenting an alternative proof that probabilistic conditional independencies do not have a complete axiomatization. More importantly, this approach may lead to a complete axiomatization of a subclass of probabilistic conditional independencies, which is crucial to qualitative reasoning.